%% file: main.tex
\pdfoutput=1
\documentclass[conference]{IEEEtran}

\IEEEoverridecommandlockouts                              %

\usepackage{graphicx} %
\usepackage{amsmath}
\usepackage{amssymb}
\usepackage{amsfonts}
\usepackage{booktabs} %
\usepackage{color}

\usepackage[table]{xcolor}
\usepackage{cite}
\usepackage{url}
\title{Contact-Guided Exploration for Non-Prehensile Locomanipulation with Multi-Critic RL}

\author{Simone Tolomei$^*$, Mayank Mittal$^{\dagger\ddagger}$, Franco Angelini$^*$, Manolo Garabini$^*$, Paolo Salaris$^*$, Marco Hutter$^\dagger$%
\thanks{$^{*}$Centro di Ricerca E. Piaggio, Dipartimento di Ingegneria dell'Informazione, Universit\`{a} di Pisa, Italy; $^\dagger$ETH Z\"{u}rich, Switzerland; $^{\ddagger}$NVIDIA.}%
\thanks{Email: {\tt\small simone.tolomei@phd.unipi.it}}
\thanks{This work was supported in part by the Next Generation EU Project
“Ecosistema dell’Innovazione” Tuscany Health Ecosystem (THE, PNRR,
Spoke 9: Robotics and Automation for Health) under Grant ECS00000017,
in part by the Italian Ministry of University and Research (MUR)—Fondo
Italiano per la Scienza Applicata (FISA) through the OCCAM Project CUP
I53C25000700001 under Grant FISA 2023-00324, and in part by MUR in
the framework of the ``CrossLab'' and ``FoReLab'' (Future-oriented Research Lab) Projects, and in part by the Swiss National Science Foundation through the National Centre of Competence in Automation (NCCR automation).  This project has received funding from the European Union’s Horizon Europe Framework Programme under grant agreement No 101121321. }%
}

\begin{document}
\bstctlcite{IEEEexample:BSTcontrol}

\maketitle

\begin{abstract}
Non-prehensile manipulation offers versatile skills for moving and rearranging heavy or bulky objects, particularly when combined with a mobile manipulation platform. 
However, both model-based and model-free approaches struggle with the complex hybrid dynamics and the sparsity of the contact in these tasks.
To address these challenges, we propose a contact-guided exploration strategy implemented within a Multi-Critic Reinforcement Learning (RL) framework. A dedicated exploration critic is trained with a dense contact-seeking reward that guides the end-effector toward meaningful contact points; its influence is progressively decayed to recover a task-optimal policy.
We obtain candidate interaction points from a general-purpose grasping algorithm, enabling the exploration mechanism to generalise across various object geometries.
We evaluate the approach on multiple tasks, including box pushing, chair transportation, and a dishwasher opening task.
Finally, we validate the chair transportation policy through extensive experiments on a quadrupedal mobile manipulator, demonstrating deployable non-prehensile manipulation in the real world.
Video and project page: \url{https://tolomeis.github.io/contact-guided-exp/}.
\end{abstract}


\section{Introduction}
\label{sec:intro}

\input{S1_introduction}

\section{Related Works}
\label{sec:related_work}
\input{S2_SoA}

\section{Methodology}
\label{sec:method}
\input{S3_Method}

\input{S4b_results.tex}

\section{Conclusion}
\label{sec:discussion}

This paper presented a Multi-Critic RL framework designed to overcome standard reinforcement learning's tendency to converge to sub-optimal local minima in complex manipulation tasks. 
By decoupling value function approximations for exploration and task performance, the proposed architecture encourages contact-rich interactions during early training while gradually shifting focus to task objectives.
Experimental results across diverse simulated environments and real-world hardware trials demonstrate that our approach is robust to external perturbations and generalizes to novel objects and payloads. 
Notably, the system successfully manipulated objects exceeding the robotic arm's rated static payload, confirming that the learned policies effectively leverage contact dynamics and whole-body control for heavy-duty transport.

While evaluation spans three distinct tasks, further testing across broader domain shifts is required to fully demonstrate robustness; additionally, relying on external motion capture limits field deployment.
Additionally, replacing fixed critic weight schedules with an adaptive performance-based curriculum may reduce hyperparameter sensitivity and improve training autonomy.

\bibliographystyle{IEEEtran}
\bibliography{IEEEabrv, refs}

\end{document}

%% file: S1_introduction.tex
Mobile manipulation is a fundamental skill for robots operating in unstructured human environments, such as households, construction sites, and warehouses. Traditional pick-and-place approaches \cite{mahler2019learning} are fundamentally limited by gripper size and payload capacity, making non-prehensile skills, such as pushing, pulling, or sliding, a powerful alternative for manipulating heavy or bulky objects \cite{lynch_stable_1996}.
Controlling these interactions, however, is notoriously difficult due to the hybrid nature of the dynamics. 
Unlike grasping, where the connection between the robot and object is rigid, non-prehensile manipulation relies on unilateral contact constraints and friction cones. 
This challenge is significantly amplified in locomanipulation due to additional constraints. The system must simultaneously govern the floating base's stability, the manipulator's workspace limits, and the object's dynamics. 
Traditional model-based approaches, such as Model Predictive Control (MPC), have shown success in this domain \cite{chiu2022collision, mittal2022articulated}. 
However, these methods typically rely on simplified physical models to maintain computational tractability \cite{sleiman2023versatile, brudigam2024jacta}, limiting their ability to capture discontinuous contact dynamics, handle model-object mismatch, and replan in real-time when contact is lost or slips occur.

\begin{figure}[!t]
    \centering
    \includegraphics[width=\linewidth]{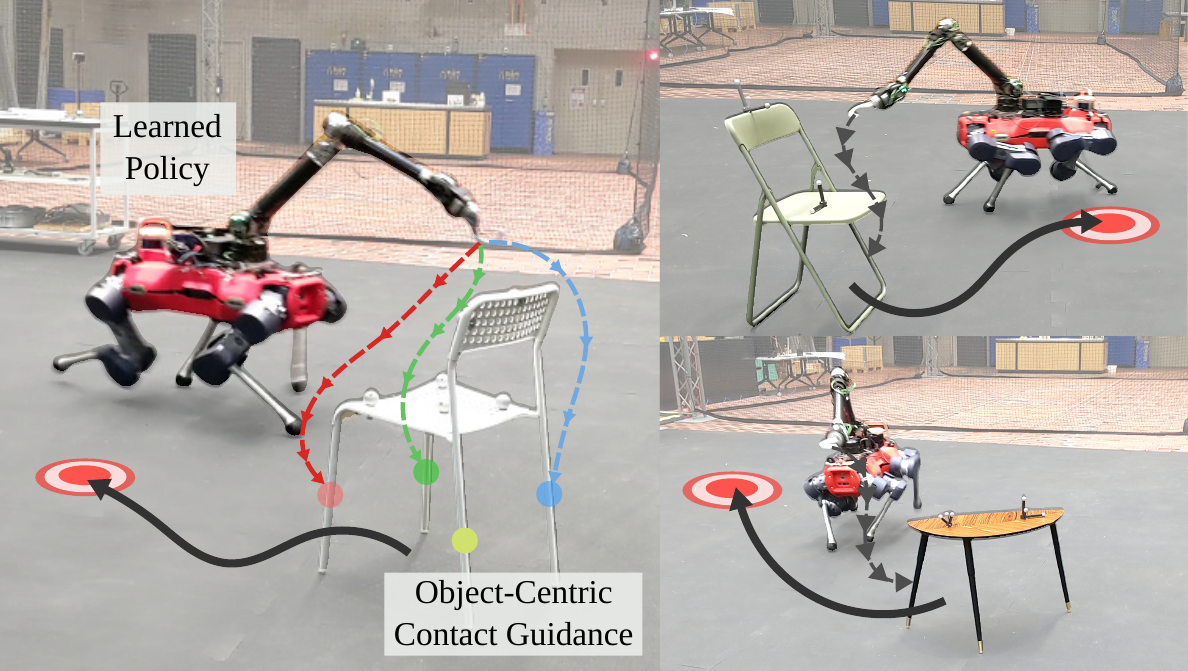}
    \vspace{-20pt}
    \caption{\textbf{Learning Non-Prehensile Manipulation via Contact-guided exploration. } We guide reinforcement learning exploration using object-centric contact priors. The colored dashed lines illustrate the learned policy’s ability to execute contact-rich behaviours (e.g., pushing a chair to a target) by focusing on high-probability interaction regions.}
    \label{fig:coverpic}
    \vspace{-16pt}
\end{figure}

Deep Reinforcement Learning (DRL) offers a robust, model-free alternative that can learn policies for contact-rich tasks directly from experience \cite{jeon2023learning}. 
However, while these methods have achieved notable success in contact-rich tasks such as in-hand manipulation \cite{andrychowicz2020learning}, applying them to long-horizon tasks such as locomanipulation remains significantly challenging. 
Sparse exploration is a common issue in manipulation, where, unless the policy discovers a valid contact and applies force in the right direction, it receives no positive feedback. 
This problem is amplified in locomanipulation, due to base-arm coordination and the high number of degrees of freedom.
As a result, the reward signal is dominated by regularisation terms that penalise abrupt motions or high energy consumption. This imbalance frequently drives the agent into a local minimum where it completely avoids contacts to minimise penalties, resulting in a fragile trade-off that typically necessitates extensive trial-and-error.
While recent works have demonstrated successful whole-body manipulation for geometries like boxes or cylinders \cite{dadiotis2025dynamic}, these methods rely on large, inherently stable, contact surfaces and guide exploration through points uniformly sampled on the object.
We extend this idea to more generic geometries, where uniform sampling often yields kinematically or physically infeasible contacts on non-convex objects, and we show that a multi-critic formulation achieves comparable exploration guidance without requiring constraint-based optimization.

We address the exploration bottleneck by proposing a strategy that guides the learning process from exploration to robust task execution. 
We leverage a general-purpose grasping algorithm to identify candidate interaction regions, such as chair legs or dishwasher handles, and synthesize a dense exploration reward that encourages the end-effector to reach these points. 
We implement this weight schedule using a multi-critic RL architecture. 
This allows us to assign a dedicated critic to the exploration objective, the influence of which is dynamically decayed during training. 
This forces the policy to transition from guided contact synthesis to a robust, task-optimal policy that respects physical constraints.

We validate our approach on two distinct tasks: non-prehensile chair transportation and box pushing. We demonstrate that our method significantly outperforms standard scalar-reward baselines and fixed-weight Multi-Critic approaches. Finally, we validate the chair transportation policy through extensive experiments (Fig.~\ref{fig:coverpic}) demonstrating zero-shot generalization to real-world objects with varying geometries and masses.

In summary, the main contributions of this paper are:
\begin{itemize}
    \item A contact-guided exploration strategy for non-prehensile interaction, implemented as a decay schedule over the weight of a dedicated exploration critic head in the advantage mixing, which guides the end-effector toward meaningful contact points and is progressively phased out to recover a task-optimal policy.
    \item A systematic evaluation across the box pushing and chair transportation tasks, and a further qualitative evaluation on dishwasher opening.
    \item Hardware validation on chair transportation with the ALMA quadrupedal mobile manipulator, demonstrating deployable behaviours and robustness to varying object geometries and masses.
\end{itemize}

%% file: S2_SoA.tex
\subsection{Non-Prehensile Manipulation}
Non-prehensile actions such as pushing, pivoting, and sliding allow robots to manipulate objects that are too heavy to lift or too large to grasp \cite{lynch_stable_1996}.
Early approaches relied heavily on model-based planning and analytical mechanics to determine contact modes and stability constraints \cite{pang2023global}. While effective for known, convex objects, these methods struggle with the uncertainty inherent in real-world friction and complex object geometries. 
Recent works have extended these concepts to mobile manipulators, utilising Model Predictive Control (MPC) to optimise whole-body motions \cite{chiu2022collision, mittal2022articulated}. However, these optimisation-based frameworks often require precise mesh models and are computationally expensive to re-plan in real-time when contact is lost or slips occur.

\subsection{RL for Locomanipulation}
Deep Reinforcement Learning (DRL) has emerged as a robust alternative for contact-rich tasks \cite{cheng2023legs, jeon2023learning}, 
learning complex behaviors directly from experience, or in combination with model-based approaches \cite{cheng2025rambo}.

In locomanipulation, RL has successfully solved tasks such as opening articulated doors and dishwashers \cite{sleiman2024guided} or pushing large obstacles \cite{dao2024sim, dadiotis2025dynamic}.
A key challenge in these approaches is the sparse reward problem; random exploration rarely leads to successful manipulation of complex objects. 
To avoid this, many methods rely on expert demonstrations \cite{xu2025intermimic, ross2011reduction, rajeswaran2017learning, nair2018overcoming} or teleoperation data \cite{fu2024mobile, chi2025diffusion}, while others employ visual pre-training \cite{liu2024visual, he2024learning}. 
However, obtaining high-quality motion capture data or expert teleoperation for dynamic non-prehensile interactions is difficult and not easily scalable \cite{brudigam2024jacta}. 
In contrast, our work learns from scratch using object-centric contact priors as an exploration bias,  and gradually decays their influence to recover an optimal policy without assuming a specific morphology.

\subsection{Multi-Critic and Curriculum Learning}
Balancing exploration and exploitation is a fundamental challenge in RL. 
In tasks with conflicting objectives, such as approaching an object (exploration) versus moving it precisely (task), standard scalar reward function often leads to suboptimal local minima.
Multi-Critic architectures \cite{vijayan2025multi, mysore2022multi, cheng2023multi}
address this by learning separate value functions for different reward components. This decomposition allows for more stable gradient estimation and enables the weighting of objectives to be adjusted dynamically during training. 
We leverage this architecture to implement a decay schedule: we emphasise exploration critics early in training, and transition to task-specific critics to refine the manipulation strategy.

%% file: S3_Method.tex
We address non-prehensile locomanipulation tasks where a quadrupedal mobile manipulator must move objects, including chairs and boxes, to a target position. The core challenge across all tasks is computing whole-body references that successfully manipulate the object while preventing tipping and keeping the robot within its kinematic and dynamic limits.

\begin{figure*}[t]
    \centering
    \includegraphics[width=\linewidth]{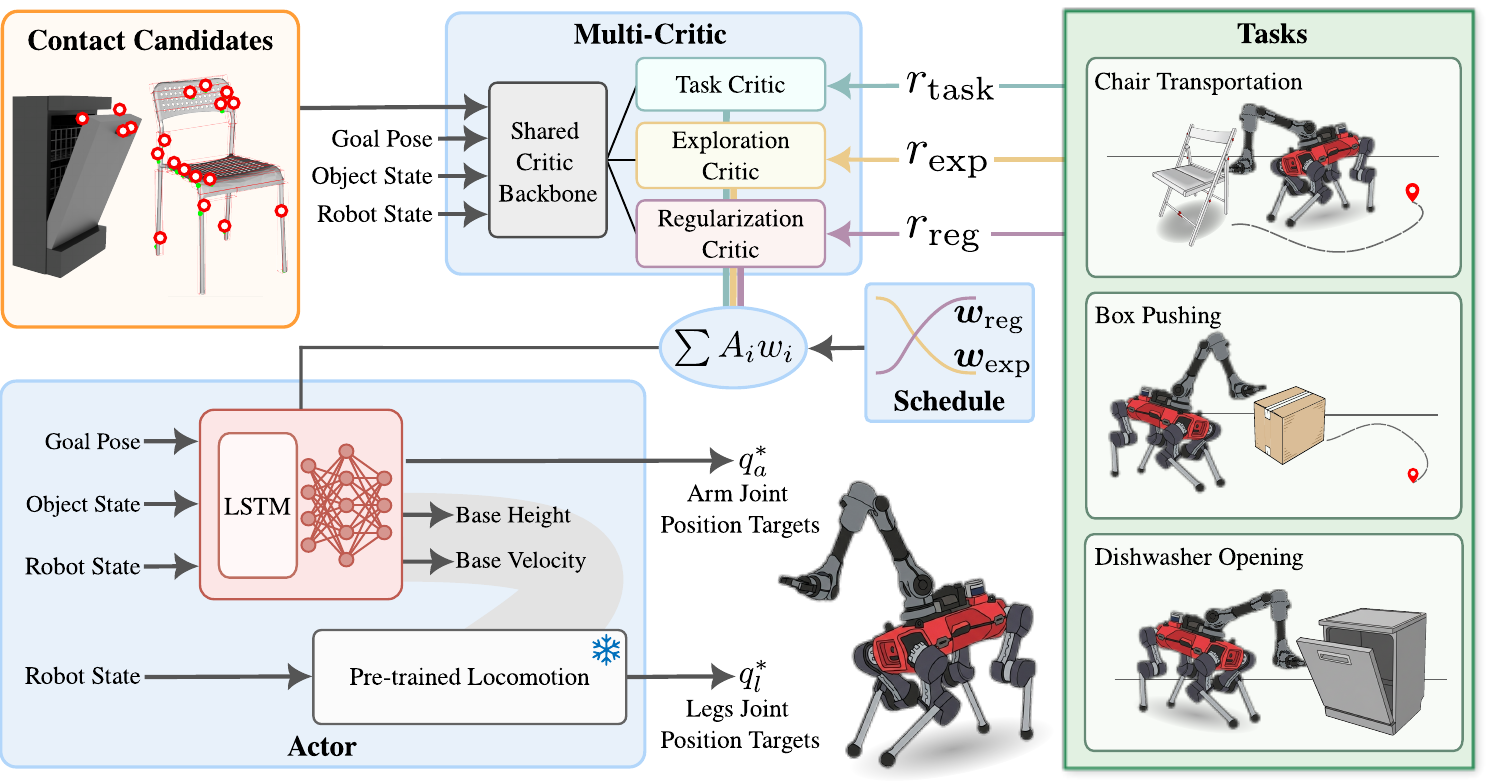}
    \vspace{-20pt}
    \caption{\textbf{Overview of the System Architecture.} (Top Left) Contact candidates are generated from object meshes using surface sampling or grasping algorithm to guide exploration. (Center) The Multi-Critic module employs a shared backbone with three specialized heads for task ($r_{\text{task}}$), exploration ($r_{\text{exp}}$), and regularization ($r_{\text{reg}}$) rewards. These are combined using a schedule-weighted advantage $\sum A_i w_i$, where the influence of exploration ($w_{\text{exp}}$) and regularization ($w_{\text{reg}}$) is decayed according to a predefined schedule. (Bottom Left) The Actor utilizes an LSTM-based policy to output target arm joint positions $q_a^*$, base height, and base velocity. The base commands are realized by a frozen, pre-trained locomotion policy that generates low-level leg joint targets $q_l^*$. (Right) The versatility of the architecture is demonstrated across three non-prehensile tasks: chair transportation, box pushing, and dishwasher opening.}
    \label{fig:diagram}
    \vspace{-10pt}
\end{figure*}

The controller follows a hierarchical approach (Fig.~\ref{fig:diagram}). A high-level
policy outputs target arm joint positions $q^{*}_{a}$, a target base SE(2) velocity
($v^{*}_{x}, v^{*}_{y}, \omega^{*}_{z}$), and a target base height $h^{*}$.
Active control of the base height is necessary to mitigate shoulder joint limits. Specifically, when the arm is extended forward to interact with objects near the ground, the shoulder joint reaches its limit, triggering the failsafe protection. By modulating $h^{*}$, the policy maintains the arm within a feasible kinematic configuration. 
The base commands are processed by a pre-trained locomotion policy, which tracks them and outputs target joint position targets for the legs. 
This policy, similar to~\cite{miki2024learning}, tracks a 4-dimensional base command $(v_{\textnormal{x}}, v_{\textnormal{y}}, w, h)$, which includes the $SE(2)$ twist and the base height $h$.
The policy receives the arm joint positions as part of its observations and is trained in simulation under randomised arm motions spanning the entire arm workspace.

\subsection{Exploration and Contact Sampling}
To address the sparse reward nature of non-prehensile manipulation, we introduce a dense exploration reward $r_{\text{exp}}$ that guides the end-effector toward meaningful interaction regions on the object.
At each episode reset, we sample a 3D target contact point $\mathbf{p}_{target}$ from a set of candidate points generated by a grasping algorithm, and reward the policy for reducing the end-effector distance to this target.
This exploration objective is not enforced through a fixed shaping term: instead, we treat it as a dedicated reward group with its own critic head, and we progressively decay its weight during training. This decouples contact-seeking exploration from the final task objective, enabling a smooth transition from ``find contact'' to ``solve the task'' without locking the policy into the specific suboptimal behaviour of exploration seeking.

\begin{figure}
    \centering
    \includegraphics[width=0.9\linewidth]{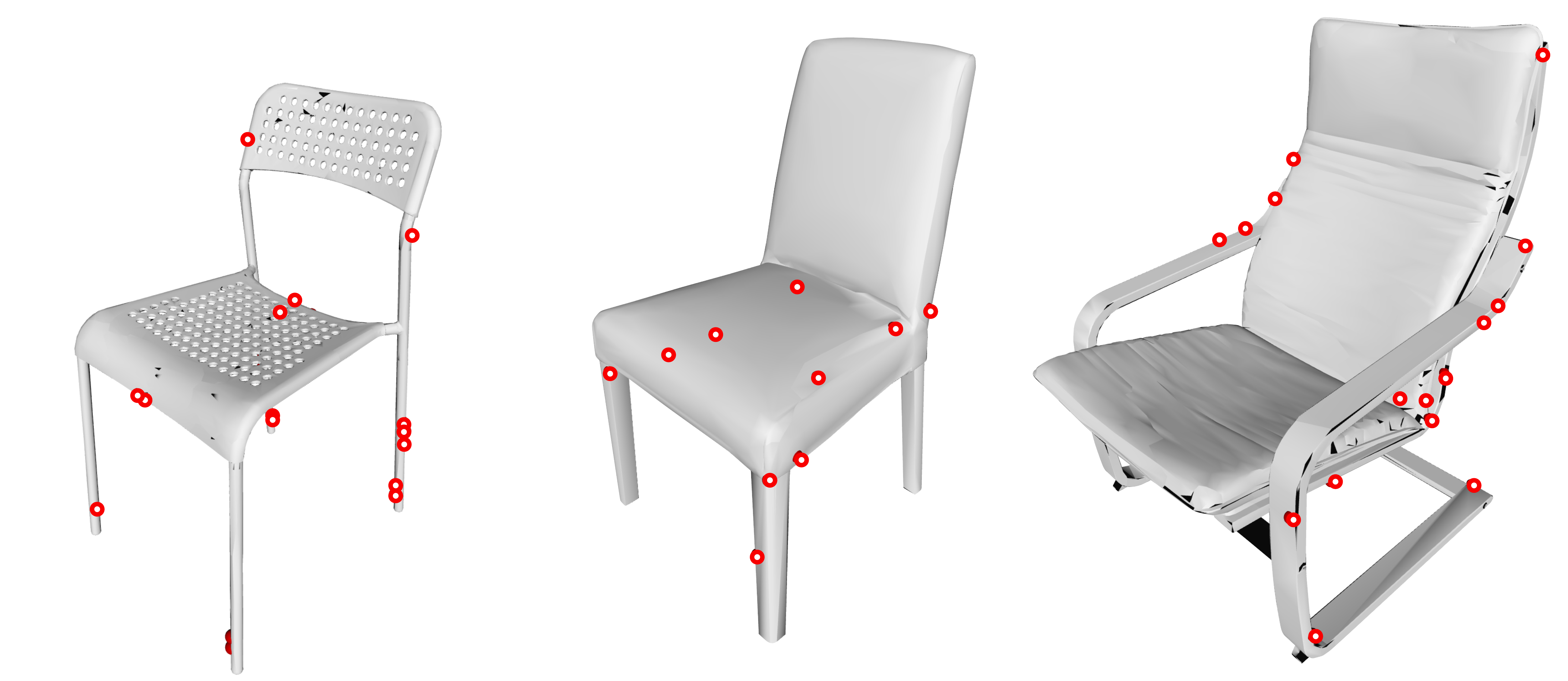}
    \vspace{-8pt}
    \caption{\textbf{Contact Candidates on Complex Geometries.} Red markers indicate candidate contact points computed via a grasping algorithm. During training, a single candidate point is sampled to serve as a target for the end-effector position. The dataset includes both procedurally randomized chair models and a curated subset of CAD models from the IKEA furniture dataset.}
    \label{fig:grasp_point}
    \vspace{-10pt}
\end{figure}

We adapt an off-the-shelf grasping algorithm \cite{palleschi_grasp_2023} to process the object mesh and generate a set of candidate contact points (Fig.~\ref{fig:grasp_point}). 
Using this sampling, we found that the number of candidates must strike a trade-off between diversity and sample efficiency: with too few points, the policy tends to overly track their positions, even if it comes at the cost of degraded performance on the main moving task.
With too many points, exploration effectively degenerates toward uniform mesh sampling, which greatly slows down contact discovery for complex and non-convex objects.
An intermediate number of candidates provides enough variability to avoid overfitting while keeping exploration focused on interaction‑relevant regions. 
In contrast, naive uniform surface sampling on the raw or convex‑hull mesh often yields kinematically unreachable or physically poor contact locations for highly non‑convex furniture geometries, making exploration substantially less efficient.
We sample a set 25 points returned from the algorithm, and, at each episode reset, we sample $\mathbf{p}_{target}$ from this set.
Although these points are typically optimised for grasping stability, we found them feasible as candidates for non-prehensile interactions. 
This generalization strategy enables the policy to better scale to a high number of complex shapes, where manual specification of contact points would be intractable.
For the box pushing tasks, similar to~\cite{dadiotis2025dynamic}, we uniformly sample candidate contact points across the visible surfaces rather than relying on the grasping algorithm.

\subsection{Multi-Critic RL Formulation}

We formulate the control problem using a Multi-Critic Reinforcement Learning framework
to effectively manage competing objective functions. Our approach builds upon Proximal Policy Optimization (PPO)~\cite{schulman_proximal_2017}
and extends it to support multiple value function approximations for distinct reward groups.

\paragraph{Reward Grouping}
To avoid the challenges of manual reward shaping, the environment provides a vector-valued reward $\mathbf{r}_t \in \mathbb{R}^H$, where $H=3$. We decompose the total return into three functional groups: task performance ($r_{\text{task}}$), exploration incentives ($r_{\text{exp}}$), and action regularization ($r_{\text{reg}}$).

\paragraph{Multi-Head Value Estimation and Advantage Mixing}

We define a set of value function heads $\{V_{\phi_h}\}_{h=1}^H$ to estimate the expected return for each reward stream. For each head $h$, the temporal difference (TD) error is: $\delta_{t}^{h} = r_{t}^{h} + \gamma V_{\phi_h}(s_{t+1}) - V_{\phi_h}(s_{t})$. The corresponding advantage $A_{t}^{h}$ is estimated as:
$
    A_{t}^{h} = \sum_{k=0}^{\infty}\gamma^{k} \delta_{t+k}^{h}\textnormal{,}
$
where $\gamma$ is the discount factor. 

To update the policy $\pi_\theta$, we aggregate these per-head advantages into a composite advantage $A_t$ via a weighted sum $A_{t} = \sum_{h=1}^H w_{h} A_{t}^{h}\textnormal{,}$
where $w_{h}$ represents the relative importance of reward group $h$. This formulation provides explicit control over the influence of exploration and regularization on the policy gradient, mitigating the sensitivity typically associated with a scalar reward.

\paragraph{Optimization Objective}

The policy parameters $\theta$ are optimized by maximizing the PPO clipped surrogate objective $L^{CLIP}$ \cite{schulman_proximal_2017} using the composite advantage $A_t$. Simultaneously, the collective value function loss $L^{VF}$ is minimized as the weighted sum of mean squared errors across all heads:
$
L^{VF}(\phi) = \sum_{h=1}^{H} w_{h} \left| V_{\phi_h}(s_{t}) - R_{t}^{h} \right|^2 \textnormal{,}
$
where $R_{t}^{h}$ is the discounted empirical return for head $h$.

\subsection{Training Details}

Training is performed using the Isaac Lab framework
\cite{mittal2025isaaclab}, parallelised over 4096 environments, with a simulation $dt $ of $0.005$\,s and a control $dt$ of $0.02$\,s.
Each environment contains the robot, the target object, and a randomly sampled goal
position. 
The robot's initial pose, joint states, and the object's physical properties (mass, friction) are randomised at every episodic reset to ensure robustness. We modify the PPO implementation from RSL-RL~\cite{schwarke2025rsl} for multi-critic setup.

\subsubsection{Asset Generation} For the chair transportation task, the training objects consist of a mixture of 15 chairs from the IKEA furniture dataset mixed with 100 procedurally randomised chairs. The latter are generated by sampling key geometric parameters from uniform distributions: total chair height $\in [0.70, 1.10]$\,m, seat height from ground $\in [0.35, 0.55]$\,m, seat width and depth each $\in [0.40, 0.60]$\,m, seat and backrest thickness each $\in [0.02, 0.05]$\,m, and leg cross-section $\in [0.03, 0.06]$\,m. 
This diversity in geometry, combined with the IKEA meshes, encourages the policy to generalise across a broad range of real-world chair morphologies. 

\subsubsection{Observation and Action Spaces}
The policy action space consists of a 4D base command ($v^{*}_{x}, v^{*}_{y}, \omega
^{*}_{z}, h^{*}$) and 6D target joint angles for the arm $q^{*}_{a}$. The observation
space differs between the actor and critic networks. The actor observes
proprioceptive data (joint states, projected gravity vector, base velocity) and
object state (object position, orientation, and error relative to the goal).
All quantities are expressed in the robot's base frame. 
The critic network accesses privileged information, including the object's linear and angular velocities and
the specific target contact point on the object.

\subsubsection{Command Sampling}
Similar to~\cite{dadiotis2025dynamic}, the goal command is defined as a 2D target position, expressed relative to the object's initial position. 
At each  reset, the command is sampled uniformly within a disk of radius $2$\,m centered on the object.

\subsubsection{Network Architecture}

The actor network processes inputs through an LSTM layer with a hidden dimension of $256$, followed by a Multi-Layer Perceptron (MLP) with hidden layers of $[256, 128, 64]$ and ELU activations. The final layer outputs the parameters of a Gaussian distribution for the action space. The multi-critic network employs a shared structure to promote feature extraction across objectives. 
It utilizes a common LSTM and initial MLP layers, branching only at the final output layer into three distinct heads to provide independent value function estimates for the task, exploration, and regularization reward groups. This leads to negligible differences in memory usage and wall‑clock training time under identical training conditions.

\subsubsection{Multi-Critic Weight Scheduling}

We employ a linear schedule for the critic weights $w_h$. 
The exploration weight is linearly annealed from $0.1$ to $0.01$ between the training steps $5\text{k}$ and $10\text{k}$. 
In contrast, the regularization weight increases linearly from $0.15$ to $0.24$. 
Outside this range, the weights remain constant. 
The task weight is fixed throughout training, $w_{\text{task}} = 0.75$.
We tuned this schedule on the chair‑transportation task: we begin decaying the exploration weight once the policy reliably discovers contact and complete the decay by 10k updates.
The same schedule is reused unchanged for all tasks.

We report the main hyperparameters and reward weights in Table~\ref{tab:hyperparameters}. 
We note that the Hook Height term uses a large coefficient to rescale an underlying exponential penalty  $r_{\textnormal{hk}} = e^{\max(0, 0.15 - z_{\text{ee}})} - 1$, whose raw values are otherwise several orders of magnitude smaller than the other rewards.

\begin{table}[t]
    \caption{Reward Structure and Simulation Parameters}
    \vspace{-5pt}
    
    \label{tab:hyperparameters}
    \centering
    \begin{tabular}{l c}
        \toprule \textbf{Reward Term}                                          & \textbf{Weight}       \\ \rowcolor{gray!15}
        \midrule \multicolumn{2}{c}{\textit{Task Group (Goal Achievement: $r_{\text{task}}$)}}               \\
        Object Vel. to Goal                                                              & $10.0$                          \\
        Object Pos. Tracking                                                             & $15.0$                          \\
        \textbf{Group Weight $w_{\text{task}}$}                                          & \textbf{0.75}                   \\ \rowcolor{gray!15}
        \midrule \multicolumn{2}{c}{\textit{Exploration Group (Contact Guidance: $r_{\text{exp}}$)}}      \\
        EE Tracking Contact Point                                                        & $3.0$                           \\
        \textbf{Group Weight $w_{\text{exp}}$}                                           & \textbf{0.1 $\to$ 0.01}         \\  \rowcolor{gray!15}
        \midrule \multicolumn{2}{c}{\textit{Regularization Group (Smoothness \& Safety: $r_{\text{reg}}$)}} \\
        Base Action Rate                                                                 & $2.0$                           \\
        Arm Action Rate                                                                  & $0.5$                           \\
        Hook Height ($z$)                                                                & $-5000.0$                       \\
        Action Limits                                                                    & $-1.0$                          \\
        \textbf{Group Weight $w_{\text{reg}}$}                                           & \textbf{0.15 $\to$ 0.24}        \\
        \bottomrule
        \toprule
        \textbf{Simulation and Training Parameters}                               & \textbf{Value / Range}          \\
        \midrule \rowcolor{gray!15}
        \multicolumn{2}{c}{\textit{Domain Randomization}}                                 \\
        Object Friction                                                                  & $[0.2, 1.5]$                    \\
        Object Mass                                                                      & $[2.0, 4.0]$\,kg                \\
        Robot Base Mass                                                                  & $\pm 5.0$\,kg                   \\
       \midrule \rowcolor{gray!15}
        \multicolumn{2}{c}{\textit{PPO Parameters}}                                 \\
        Actor / Critic Learning Rate                                           & $1\times10^{-4}$ / $1\times10^{-4}$ \\
        Discount Factor $\gamma$   / GAE Parameter $\lambda$                                                  & $0.99$  /         $0.95$                      \\
        Clip Range    /   KL Target                                                         & $0.2$ /  $0.01$                          \\
        PPO Epochs  / Mini-batches per Epoch                                                & $5$  / $4$                           \\
        \bottomrule
    \end{tabular}
    \vspace{-15pt}
\end{table}

%% file: S4b_results.tex
\section{RESULTS}
\label{sec:sim_results}

We first validate our approach in simulation to verify its efficacy in overcoming the exploration bottleneck. We analyze the policy's performance quantitatively against baselines and qualitatively to understand the learned manipulation strategies.

\begin{figure*}[t]
    \centering
    \includegraphics[width=\linewidth,trim={0 50 0 0},clip]{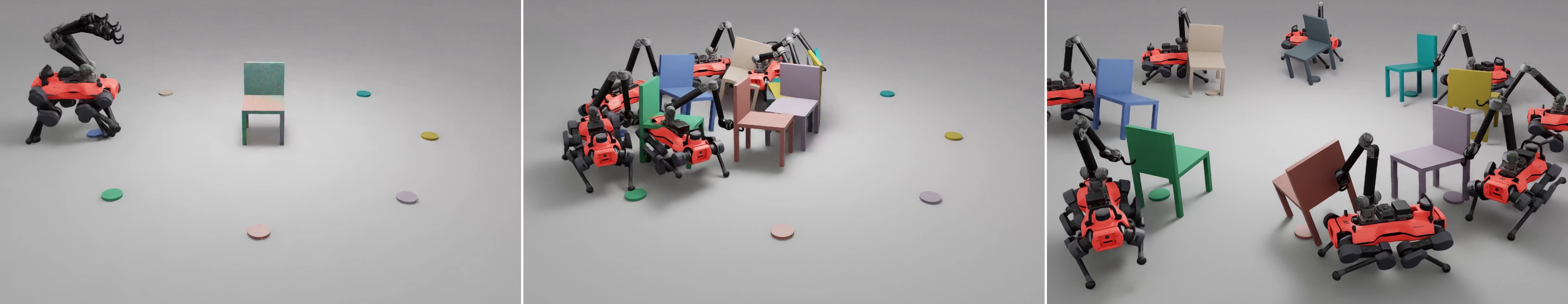}
    \vspace{-20pt}
    \caption{\textbf{Manipulation Strategy for Target-Directed Transportation.} The learned policy exhibits an orientation-aware strategy to maximize control authority over the chair. The agent is initialized relative to the object with various target locations (colored discs). The overlapping silhouettes illustrate the continuous transition from initial approach to stabilized pushing, demonstrating the policy's ability to balance base locomotion with precise arm coordination.}
    \label{fig:circleseq}
    \vspace{-10pt}
\end{figure*}

\begin{figure}
    \centering
    \includegraphics[width=1\linewidth]{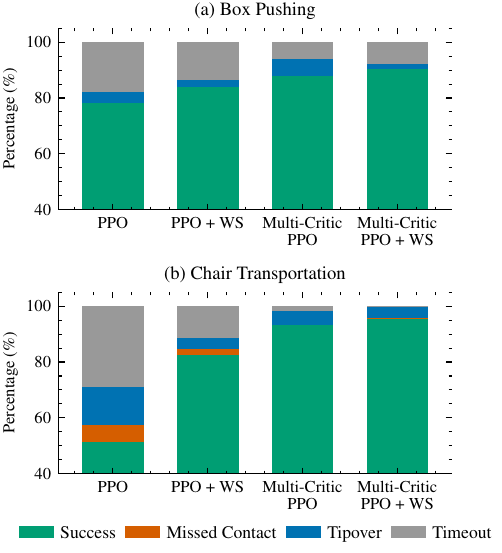}
    \vspace{-20pt}
    \caption{\textbf{Quantitative performance and failure mode analysis.} Comparison of success rates and termination causes for (a) Box Pushing and (b) Chair Transportation. All methods use the contact-guided exploration reward during training. While vanilla PPO and PPO with Weight Schedule (WS) suffer from significant rates of Timeouts and Missed Contacts, the addition of the Multi-Critic architecture substantially improves efficiency. The full proposed method (Multi-Critic PPO + WS) achieves the highest success rates (over 90\%), effectively mitigating Tipover events.}
    \label{fig:barchart_combined}
    \vspace{-10pt}
\end{figure}

\subsection{Quantitative Analysis}
\label{sec:quantitative}

We test our approach (referred to as \emph{Multi-Critic PPO + WS}) against three baselines. A standard \emph{PPO} baseline with a single critic head where the reward terms are weighted to mimic the initial configuration of our critic weights. The second is \emph{PPO + WS}, which is a standard  PPO with weight decay applied directly to the reward terms; although this is inspired by~\cite{dadiotis2025dynamic}, their approach uses a constrained RL formulation, which we parameterise here as penalties and terminations to ensure a fair comparison with our method. The final baseline is \emph{Multi-Critic PPO}, where we keep the Multi-Critic formulation but fix the group weights to $w_{\text{task}}=0.75$, $w_{\text{exp}}=0.1$, and $w_{\text{reg}}=0.15$ for the full training horizon.

We measure the \textit{Success Rate} (object reaches goal within 0.2m) and analyze specific failure modes: 
\begin{itemize}
    \item \textbf{Missed Contact:} The robot fails to interact meaningfully with the object (displacement $< 0.2$ m).  

    \item \textbf{Tipover:} The object tilts beyond 35 degrees. This value was selected to label unstable configurations and does not correspond to the exact tipping angle of every object. In practice, varying this threshold within 30–40 degrees leaves the trends in Fig.~\ref{fig:barchart_combined} unchanged.

    \item \textbf{Timeout:} The object fails to reach the goal
        within the time limit.
\end{itemize}
We aggregated the results of 5 different random seeds. The average outcomes are summarized in Fig. \ref{fig:barchart_combined}. Over seeds, our methods shown standard deviation in the success rate of $0.98\%$, while Multi‑Critic PPO without scheduling, PPO + WS, and PPO obtain 4.2\%, 1.2\%, and 9.7\%, respectively. These results highlight three key findings.
\subsubsection{The Exploration Bottleneck}
The standard \textit{PPO} baseline suffers heavily from the sparse reward
problem, and, on chair transportation, it shows a high Missed Contact rate (9.1\%). Without explicit guidance, the agent frequently converges to a local minimum where it avoids interaction to minimize energy penalties. 
Adding a weight schedule to the scalar reward (\textit{PPO + WS}) (same schedule as ours, but applied to the reward-term weights rather than critic-head weights) improves contact discovery (Missed Contact drops to 4.0\%), but the conflicting objectives within a single value function lead to unstable learning and high variance.
We additionally tested a variant without an exploration reward, which fails to discover any meaningful contact with the object; the Success Rate is 0\%.

\subsubsection{Effects of Fixed-Weight Multi-Critic}
The \textit{Multi-Critic PPO} baseline, which separates the value functions
but keeps weights fixed ($w_{\text{task}}=0.75$, $w_{\text{exp}}=0.1$, $w_{\text{reg}}=0.15$), solves the contact problem effectively. However,
because the exploration reward remains active throughout the episode, the policy
over-prioritizes the contact action at the expense of transport stability. The
agent often fails to learn a smooth behaviour, leading to a significantly
higher Tipover rate.

\begin{figure*}[t]
    \centering
    \includegraphics[width=0.94\linewidth]{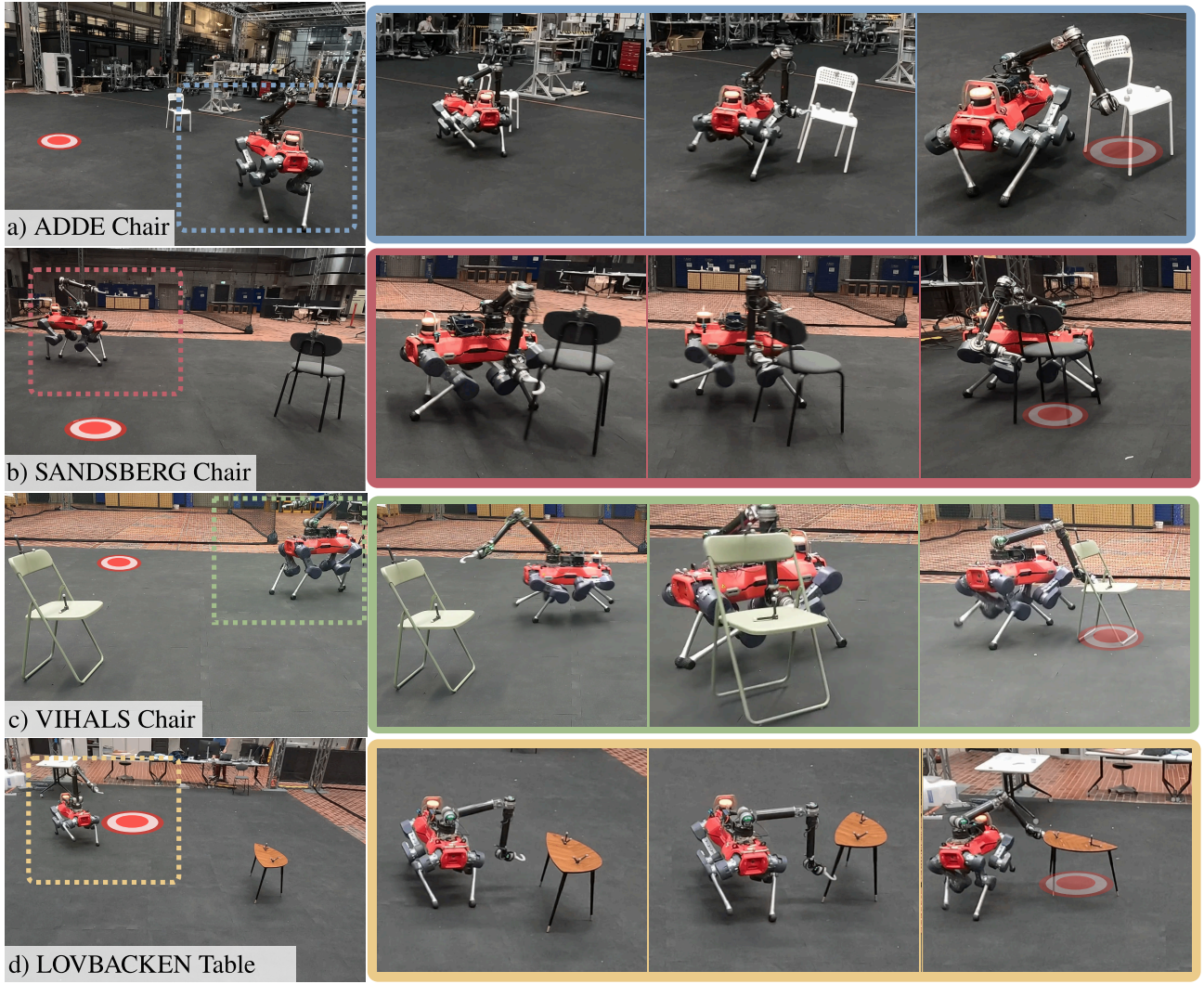}
    \vspace{-8pt}
    \caption{\textbf{Experimental validation and zero-shot real-world generalization.} We deploy the policy on a quadrupedal manipulator and evaluate it on diverse physical geometries. Panels (a)–(c) show successful transportation of multiple chair models, while panel (d) shows robustness on a LOVBACKEN three-legged table. The policy generalizes to the table’s backrest-free topology, reorienting the base to maintain an optimal contact normal and transferring contact-guided exploration from simulation to the real world without additional fine-tuning.}
    \label{fig:obj_photoseq}
    \vspace{-10pt}
\end{figure*}

\subsubsection{Benefits of Weight Scheduling}
Our proposed method achieves the highest success rate by combining the benefits of decoupled value estimation with corresponding weight scheduling. 
The initial exploration weight ensures robust contact discovery, while the decay phase forces the agent to disregard the contact point and optimize for the task physics, reducing instability and improving efficiency, achieving the best overall success rate (94.1\%) with a low tipover rate (4.4\%) and a completion time (9.2\,s) comparable to \emph{PPO + WS} (9.1\,s).

\subsection{Qualitative Behavior}
\label{sec:qualitative_sim}
Beyond success rates, we analyze the decision-making capabilities of the learned
policy. We fix the robot and object poses and sample goal positions uniformly around
the object. 
As illustrated in Fig. \ref{fig:circleseq}, the agent demonstrates an understanding of the object's topology relative to the goal. 
It consistently selects a contact point on the side aligned with the direction of motion, effectively maximizing control authority for the pulling task. Furthermore, the agent learns to position the mobile base to maintain the object within the manipulator's
optimal workspace, avoiding singularities.

\subsection{Hardware Deployment}
\label{sec:experiments}

We conducted extensive real-world experiments to evaluate the policy's transferability to physical hardware, its robustness to external disturbances, and its generalization to unseen object geometries.
The experiments were performed on the ALMA quadrupedal manipulator. 
The robot relies on on-board proprioception for base and arm control, while the object pose is tracked via an external motion capture system.

\subsubsection{Object Generalization}

\begin{figure}
    \centering
    \includegraphics[width=0.95\linewidth]{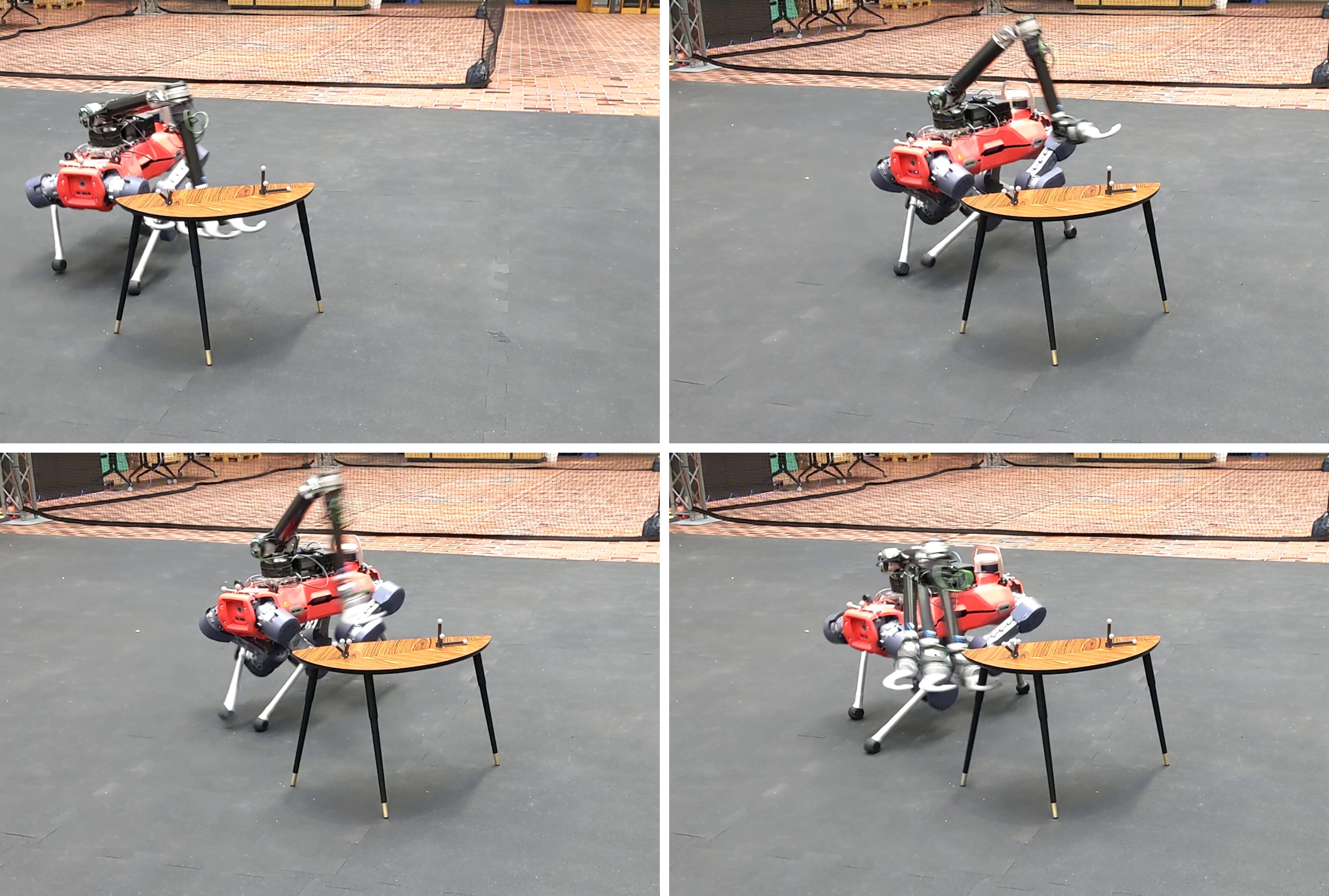}
    \vspace*{-10pt}
    \caption{\textbf{Reactive recovery behavior under contact failure.}
    Following an initial slip or failed contact attempt, the agent does not enter an unstable state but instead executes a recovery sequence: first re-positioning the base, then increasing the arm's swing amplitude and re-establishing a stable contact point to resume the transportation task.}
    \label{fig:hook_recovery}
    \vspace{-10pt}
\end{figure}

We evaluated the policy on four unseen IKEA objects (Table~\ref{tab:exps}) to test generalization, achieving an aggregate success rate of 69.0\% (40/58 trials). Benchmarking (Fig.~\ref{fig:obj_photoseq}) demonstrated consistent success on standard four-legged geometries (ADDE, SANDSBERG). To explore edge cases, we included a folding chair (VIHALS) and a three-legged table lacking a backrest (LOVBACKEN). 
Notably, the agent successfully manipulated the asymmetrical table by adapting to hook its legs, confirming the policy learns generalized non-prehensile physics rather than memorizing meshes. Furthermore, we observed an emergent recovery behavior: if an initial hook missed or slipped, the policy dynamically increased the end-effector trajectory amplitude for subsequent attempts (Fig.~\ref{fig:hook_recovery}).

\begin{table}[]
    \centering
    \caption{Hardware Experiments on IKEA Objects}
    \vspace{-5pt}
    \label{tab:exps}
    \begin{tabular}{rccc}
        \toprule
        \textbf{Object}  & \textbf{Successes} & \textbf{Total Runs} & \textbf{Success Rate} \\
        \midrule \rowcolor{gray!15}
        \multicolumn{4}{c}{\textit{Primary Benchmark}}              \\
        \textbf{ADDE}                                        & 27                 & 37                         & 72.90\%                \\
        \textbf{SANDSBERG}                                            & 8                  & 14                         & 57.14\%               \\ \midrule
        \rowcolor{gray!15}
        \multicolumn{4}{c}{\textit{Case Studies}}               \\
        \textbf{VIHALS}                                              & 3                  & 3                          & 100.00\%                 \\
        \textbf{LOVBACKEN}                                          & 2                  & 4                          & 50.00\%                  \\
        \bottomrule
    \end{tabular}
    \vspace{-16pt}
\end{table}

We found that the policy learns to command a constant target base
height, around $15$cm lower than the standard base height in locomotion, which allows
the end-effector to reach regions of the space closer to the ground without extending the arm close to the joint limit.
Despite a 94.1\% success rate in simulation, hardware success drops on ADDE (72.9\%) and SANDSBERG (57.1\%). The primary failure mode stems from lateral approaches used to avoid front self-collisions. On hardware, the resulting abrupt yaw commands degrade state estimation and induce odometry errors, leading to aggressive movements that tip the object over.

\subsubsection{Dynamic Goal Tracking}
We manually updated the goal position in real-time to simulate a moving target.
The policy successfully generalized to this dynamic scenario, adjusting the base
velocity and contact forces to steer the chair along the changing trajectory.
This confirms the policy's suitability as a local controller for high-level path
planners.

\subsubsection{High Payload Manipulation}
We attached additional weights to the chair (Fig.~\ref{fig:hw_robustness} (a)), increasing the total mass to 6.5 kg.
This exceeds the robot arm's rated static payload when fully extended. The agent
successfully completed the task, confirming that the learned non-prehensile
strategy effectively leverages the mobile base and ground support to manipulate
loads beyond the manipulator's nominal capacity.

\subsubsection{Disturbance Rejection}
\begin{figure}
    \centering
    \includegraphics[width=0.49\linewidth,trim={0 130 0 150},clip]{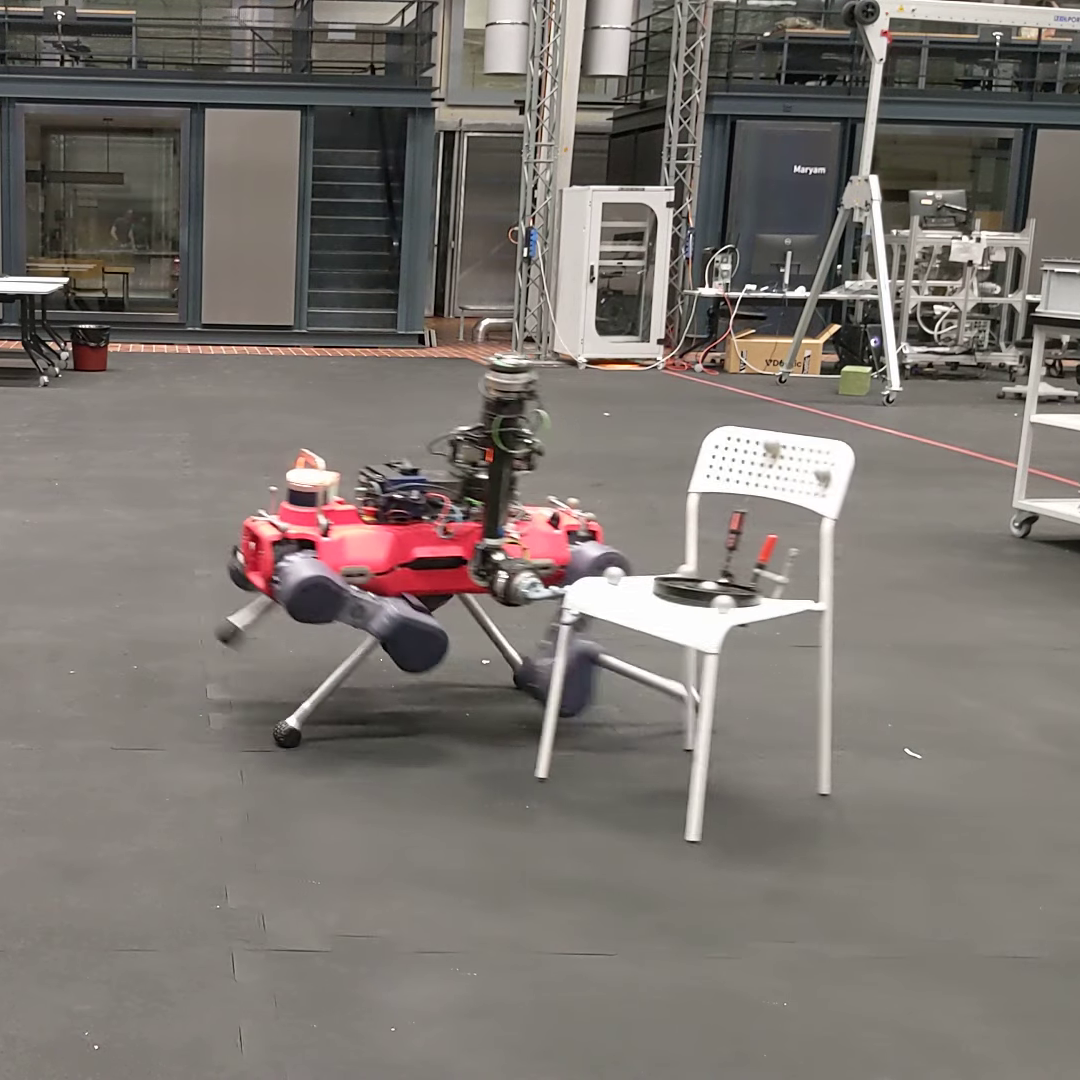}
    \includegraphics[width=0.49\linewidth,trim={0 130 0 150},clip]{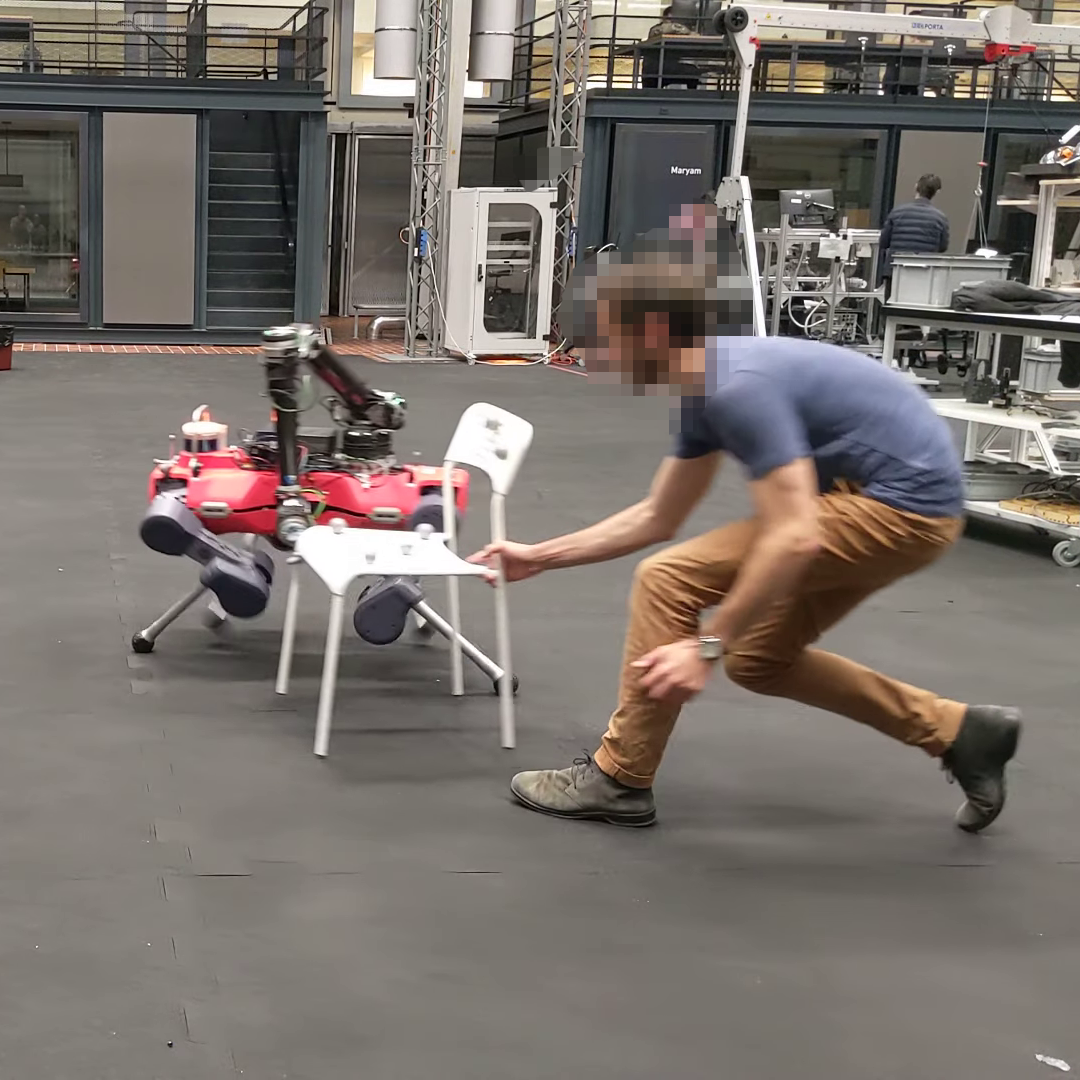}
    \vspace{-20pt}
    \caption{\textbf{Robustness to physical perturbations and payload changes.} (a) The agent transports a chair with an additional 5 kg payload, showing robustness to large, unseen variations in inertia and friction. (b) The system rejects real-time disturbances when a human abruptly shifts the object, with the policy immediately re-planning the contact trajectory.}
    \label{fig:hw_robustness}
    \vspace{-10pt}
\end{figure}

To test closed-loop stability, an operator physically pushed the chair
during transport (Fig.~\ref{fig:hw_robustness} (b)). Although trained in a static environment, the policy consistently
recovered. Following a forced disengagement, the robot autonomously re-planned its
approach, re-hooked the object, and completed the task.

\subsection{Contact Sequencing for Articulated Object Manipulation}
\label{sec:emergent_behavior}

\begin{figure}
    \centering
    \includegraphics[width=0.95\linewidth]{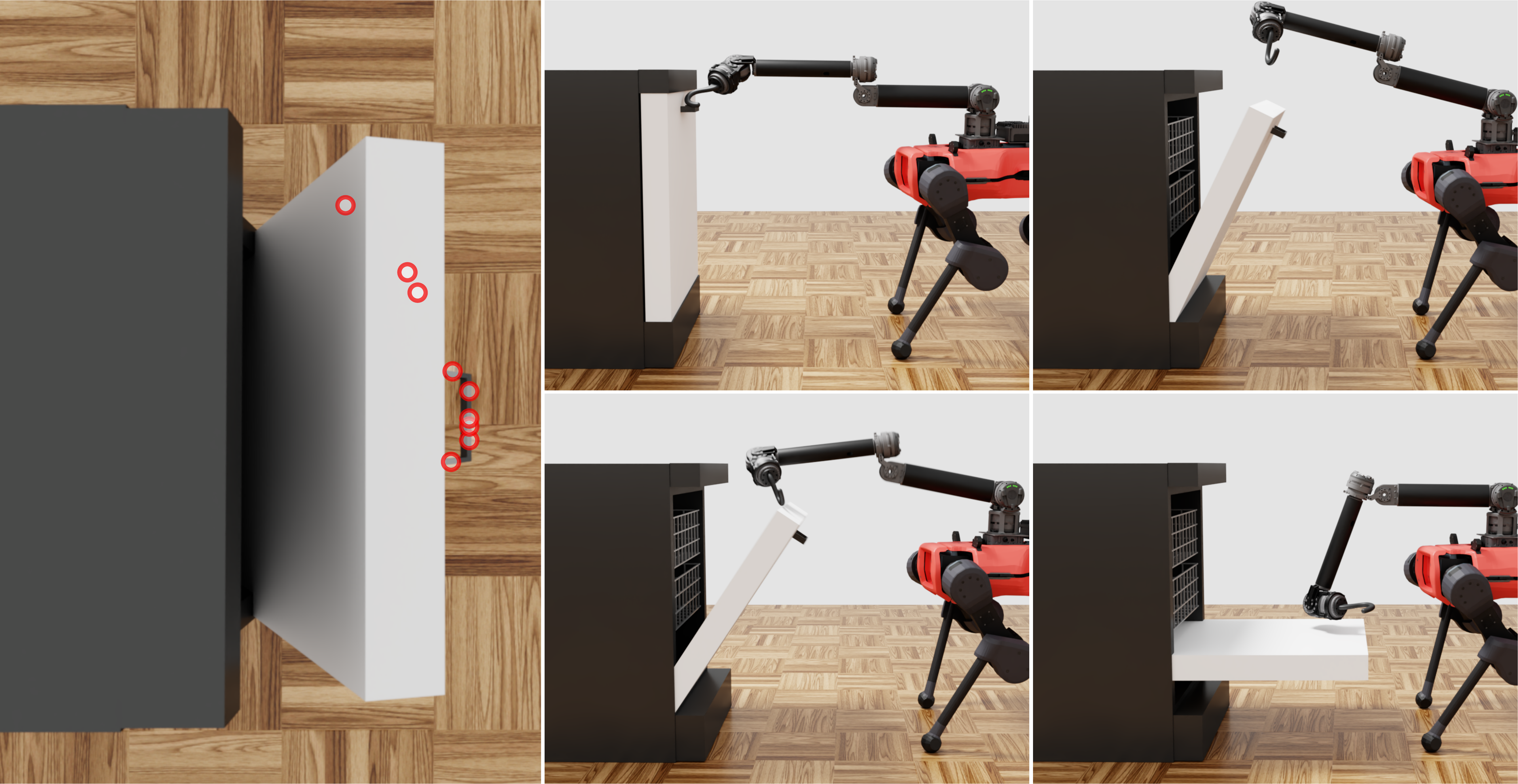}
    \vspace{-10pt}
    \caption{\textbf{Emergent contact sequencing for articulated object manipulation.} In the dishwasher opening task, the contact-guided exploration provides candidate points (red markers) on both the handle and the door panel. The learned policy demonstrates a multi-stage strategy: (Top) the agent first uses the handle to initiate the opening motion. (Bottom) As the door kinematics change, the agent switch contact to the panel, utilizing a pushing motion to lower the door.}
    \label{fig:dishwasher_seq}
    \vspace{-10pt}
\end{figure}

We further investigated our framework when interacting with articulated objects in a dishwasher-opening task.
As illustrated in Fig. \ref{fig:dishwasher_seq},
the policy first selects the handle as the optimal contact point to begin the
opening motion. As the door's orientation changes, the policy
switches its contact to the door panel to push it into a fully open horizontal
position. 
Both the handle and the panel are included in the set of contact candidates used by the exploration, and the agent learns to dynamically select and transition between them based on the evolving task dynamics.

In this task, we also observed that simple PPO reliably solves the task but typically converges to a quasi-static strategy with limited contact switching: once contact with the handle is established, the policy maintains the contact and pulls until the door opens. 
While effective in simulation, this strategy often operates close to the manipulator's joint limits. 
On average, our method reduced by $59\%$ the fraction of timesteps where the arm joint is within $10\%$ of the position limit.